\documentclass[runningheads]{llncs}

\usepackage{eccv}

\usepackage{eccvabbrv}
\usepackage{graphicx}
\usepackage{booktabs}
\usepackage[accsupp]{axessibility}
\usepackage{xcolor}

\newcommand{\revise}[1]{#1}  

\usepackage{hyperref}
\usepackage{orcidlink}

\begin{document}

\title{Can Edge-Deployable Vision--Language Models Identify Species?}

\titlerunning{Can Edge-Deployable VLMs Identify Species?}

\author{William Zhou\inst{1}\orcidlink{0009-0009-4151-380X} \and
Mayukha Siripuram\inst{2} \and
Xiao Yan\inst{3}\orcidlink{0009-0008-9484-0209} \and
Ziqi Liu\inst{3} \and
Yi Ding\inst{3}\orcidlink{0000-0002-1226-341X}}
\authorrunning{W. Zhou et al.}

\institute{Plano West Senior High School \and
Centennial High School \and
The University of Texas at Dallas}

\maketitle

\begin{abstract}
Camera traps often run in the field on edge hardware with limited or no connectivity, making small, locally-deployable vision--language models (VLMs) -- not frontier-scale ones -- the practically relevant class to evaluate for species identification. We test whether models in this deployment-relevant 2--8B range carry genuine taxonomic knowledge, evaluating four such VLMs (Qwen3-VL 2B/4B/8B, Gemma3~4B) against the domain-specific specialist BioCLIP (300M parameters) on a 96-species task, comparing clean iNaturalist photographs against camera-trap imagery from 6 LILA.science collections, on two independently-sampled evaluation sets. All models identify species far above chance, but every model -- general-purpose or specialist -- degrades sharply on field imagery (domain gaps of 9.6--26.6 percentage points, consistent across taxonomic levels and both evaluation sets), indicating the degradation reflects general image legibility rather than fine-grained discrimination failure. BioCLIP substantially outperforms every VLM tested (by 33.2--59.2 percentage points across an expanded 200-image sample for every model) despite its far smaller size, suggesting the gap reflects specialized training data rather than model scale; yet BioCLIP's own domain gap (18.0 points) is statistically indistinguishable from the best VLM's (22.3 points), suggesting the clean-to-field degradation itself is a property of the image-quality shift rather than a general-purpose-model weakness. Under open-set prompting, 5.9--9.6\% of responses are syntactically valid but taxonomically nonexistent species names; the relative fabrication-rate ranking across models replicates exactly across both evaluation sets, a more robust finding than any single point estimate.

\end{abstract}

\section{Introduction}
\label{sec:intro}

As multimodal models become ubiquitous, a natural question arises: do they contain expert-level, professional knowledge, or only everyday categories learned incidentally from broad web-scale pretraining? Species identification is an unusually clean way to test this, because the correct answer is objectively verifiable and the gap between casual and expert-level knowledge is sharp -- the difference between recognizing ``a bird'' and recognizing \emph{Bucorvus leadbeateri}. We deliberately scope this question to models in the 2--8B parameter range. This is not a compromise forced by available compute: camera traps are typically deployed in the field on edge hardware, often without reliable connectivity to query a frontier-scale API model, so 2--8B is the class of model that would actually run on-device\revise{ -- concretely, our own measured footprint (\cref{tab:latency}) puts Qwen3-VL~2B at 4.1GB, Qwen3-VL~4B at 5.6GB, and Qwen3-VL~8B at 7.2GB. All three fit within the practical capacity of an \textbf{NVIDIA Jetson Orin Nano} (8GB unified memory), which supports roughly 7--10B-parameter models at 4-bit quantization; the \textbf{AGX Orin} (64GB) extends this to roughly 70B, well beyond the range we test, and both are hardware already used in field camera-trap units~\cite{yan2025arewetherewet,vuilliomenet2026futureedgeai}. A \textbf{Raspberry Pi 5} has no CUDA-capable GPU at any RAM tier (1--16GB); at 16GB RAM, though, its practical capacity is roughly 14B parameters at 4-bit quantization (roughly 8B at 8-bit) -- enough to fit all four of our VLMs by memory alone, so this is a throughput limitation from CPU-only inference (the same CPU-spillover regime the footnote below documents), not a capacity one. This is a meaningfully more permissive picture than power-constrained phone-class hardware: In our previous work~\cite{yan2025arewetherewet}, we find only sub-4B LLMs run successfully on power-constrained \emph{mobile} hardware, with $>$30s latency even then -- single-board computers like the Raspberry Pi and Jetson family have a more generous sustained power budget than a phone, which is reflected in their larger practical capacity. We do not claim current camera-trap hardware ships with 2--8B models today; this is the range for which on-device deployment is a near-term engineering question, not a data-center dependency.} Whether a 32B or larger model behaves the same way is a genuinely open question we do not address here.

This motivates three questions, all scoped to models in this deployment-relevant range: (RQ1) how well do such VLMs perform relative to a domain-specific specialist? (RQ2) how does accuracy differ across model family, prompting style, and image treatment within this range? (RQ3) what governs the accuracy achieved -- taxonomic difficulty, and how much does image quality (clean vs.\ field-captured) matter? A methodological concern threads through all three: findings from a single sampled evaluation set can reflect the particular species sampled rather than a general property of the models. We address this by evaluating on two independently-composed sets and reporting, for every major finding, whether it replicates.

Our contribution is a joint comparison that, to our knowledge, no prior work has performed: edge-deployable general-purpose VLMs and a domain specialist (BioCLIP), evaluated identically across clean and field-degraded imagery, with every headline finding checked for replication across two independent samples. We find that VLMs in this range possess substantial taxonomic knowledge but fall well short of specialist performance -- a gap plausibly attributable to specialized training data rather than raw scale, since BioCLIP achieves this while itself being an order of magnitude smaller (300M parameters) than the largest VLM we test; that the clean-to-field accuracy drop is a property of image degradation rather than a general-purpose-model-specific weakness (since BioCLIP degrades by a statistically indistinguishable amount); and that under open-ended querying, models can confidently generate taxonomically nonexistent species names, with a fabrication-rate ranking across models that is more robust than any single accuracy number.

\textbf{Related work.} BioCLIP~\cite{stevens2024bioclip} established that a specialist model trained on curated biological imagery (TreeOfLife-10M) substantially outperforms general-purpose vision baselines, but its evaluation is confined to clean photographs, leaving open whether this advantage persists on camera-trap imagery -- the comparison at the center of our RQ1. Camera-trap-specific adaptations exist: Fabian~et al.~\cite{fabian2023multimodal} and Yang~et al.~\cite{yang2025enhancing} fine-tune models for field imagery but do not quantify image-quality characteristics, leaving unclear whether errors trace to hard taxa or hard images (RQ3); Dussert~et al.~\cite{dussert2025zeroshot} apply VLMs zero-shot to a related task (behavior, not species classification) with day/night context, but do not analyze accuracy against image quality directly, nor scope their evaluation to edge-deployable model sizes. Closest to our own analysis, Nanduri~et al.~\cite{Nanduri2026IntraAfricanGD} compare specialist and retrieval-based models across image-quality conditions, but do not include general-purpose VLMs alongside them -- the direct comparison our work provides, within the deployment-relevant size range.

\section{Methods}
\label{sec:methods}

\textbf{Dataset.} Our candidate pool spans 5{,}554 camera-trap images across 96 species, drawn from 6 LILA.science collections~\cite{lila_cct,lila_ammonitor,lila_nkhotakota,lila_serengeti,lila_idaho,lila_oregon} (a 7th, California Small Animals, was dropped after failing quality control almost entirely), plus a species-matched pool of iNaturalist~\cite{inaturalist} photographs as a clean-domain reference. Actual evaluation samples drawn from this pool are much smaller and are specified per evaluation set below; every model is tested on the same images within a given comparison. We use \emph{domain} strictly to mean image source -- clean vs.\ trap -- distinct from a model being trained for a subject-matter specialty. Bounding boxes for both domains come from MegaDetector (63.1\% success rate on iNaturalist, 18{,}169 of 28{,}790 checked)\revise{; for the trap domain, 74.9\% of the final 5{,}554-image pool used a MegaDetector-sourced box and 25.1\% a human-drawn fallback, unevenly by collection (100\% MegaDetector for Snapshot Serengeti, AMMonitor, Idaho, and CCT; 0\% for Oregon Critters, which ships with annotator-drawn boxes; 65.6\% for Nkhotakota)}.

We evaluate on two evaluation sets for two purposes. The \emph{broad} set draws 100 images/domain at random from the pool; because there are 96 species but only 100 draws with uneven per-species availability, only 55 species end up represented (this is incidental, not deliberate, and turns into a useful check: see \cref{sec:hallucination}). The \emph{focus} set instead deliberately samples 20 images/domain for each of 18 species with sufficient availability in both domains (360 images/domain), enabling per-species comparison the broad set's thin coverage cannot support. The full 96-species candidate list is shown in every multiple-choice prompt regardless of set, keeping the 1\% chance baseline identical in both. We additionally collected 100 new, non-overlapping broad-set images/domain for every model (verified zero overlap with the original samples), bringing the broad-set headline comparisons to 200 images/domain per model.

\textbf{Experimental design.} We evaluate Qwen3-VL~2B/4B/8B and Gemma3~4B (all Q4-quantized, as would be typical for edge deployment) via local Ollama (temperature~0, fixed seed, 8{,}192-token context, reasoning disabled), and BioCLIP (300M parameters) via its CustomLabelsClassifier. We compare \emph{closed-set} (multiple-choice, all 96 candidates) and \emph{open-set} (bare binomial, no candidate list) prompting for the VLMs (BioCLIP is forced-choice only, no open-set analogue), across three image treatments: \emph{cropped} (bbox + 15\% margin), \emph{original} (full frame), and \emph{boxed} (full frame, box overlaid).\footnote{The 8B model (6.1\,GB) exceeds 4\,GB consumer laptop VRAM and spills to CPU (13--96\,s/call vs.\ 1--40\,s for smaller models) -- itself a relevant data point for edge deployment, where 8B may already be impractical on the lowest-power hardware. \revise{A dedicated benchmark (\cref{tab:latency}) found 4B also exceeds this ceiling (5.6GB); only 2B is close to fitting (4.1GB).}}

\textbf{Scoring.} Correctness is scored independently at species-, genus-, and family-level (genus/family evaluated whenever identifiable even if species match fails). Under open-set prompting, every scoreable response is validated against the GBIF taxonomic backbone~\cite{gbif} and classified as a real species (wrong ID), a real genus with an invalid species epithet, or an entirely fabricated genus.

\section{Results}
\label{sec:results}

\begin{table}[tb]
\caption{BioCLIP vs.\ four edge-deployable VLMs (2--8B), clean-domain images, multiple-choice prompting, 200-image sample for every model (100 original + 100 newly-collected, verified non-overlapping).}
\label{tab:bioclip}
\centering
\small
\begin{tabular}{@{}lcccc@{}}
\toprule
Model (params) & Cropped & Original & Boxed & Pooled (BioCLIP advantage) \\
\midrule
\textbf{BioCLIP (300M)} & \textbf{89.0\%} & \textbf{94.0\%} & \textbf{86.0\%} & \textbf{89.7\%} \\
Qwen3-VL 8B & 54.5\% & 56.5\% & 58.5\% & 56.5\% (33.2 pts) \\
Qwen3-VL 2B & 39.0\% & 44.0\% & 34.5\% & 39.2\% (50.5 pts) \\
Qwen3-VL 4B & 33.0\% & 35.0\% & 42.0\% & 36.7\% (53.0 pts) \\
Gemma3 4B & 29.0\% & 32.5\% & 30.0\% & 30.5\% (59.2 pts) \\
\bottomrule
\end{tabular}
\end{table}

\textbf{RQ1: edge-deployable VLMs vs.\ a specialist.} All four VLMs identify species far above the 1\% chance rate, confirming real taxonomic knowledge in this size range -- but BioCLIP substantially outperforms every VLM tested, by 33.2 points over the best (Qwen3-VL~8B) to 59.2 over the weakest (\cref{tab:bioclip}), holding in both domains \revise{(trap, cropped treatment -- the only treatment BioCLIP was run on in this domain, so we restrict the VLM side of this specific comparison to the same treatment rather than \cref{tab:domain-gap}'s pooled figure: 71.0\% BioCLIP vs.\ 32.0\% best VLM)}. Notably, BioCLIP achieves this despite being roughly 7--20$\times$ smaller than the VLMs we test (300M vs.\ 2--8B parameters), suggesting this advantage reflects BioCLIP's specialized training data rather than model scale -- a stronger, scale-independent claim than a size-matched comparison alone would support, and one that does not require testing larger VLMs to establish. Edge-deployable general-purpose VLMs do not yet substitute for a specialist at any scale or family we tested within this range.

\revise{Part of this gap is not purely taxonomic: BioCLIP, forced-choice, always returns an answer by construction, whereas generative VLMs can refuse, hedge, or answer off-list. This rate is uneven: on trap multiple-choice prompts, Qwen3-VL~2B fails to answer 48.7\% of items, versus 1.2--3.0\% for 4B/8B and 0\% (3.7--4.3\% off-list) for Gemma3~4B. Non-answers score as incorrect throughout, inflating BioCLIP's advantage over weaker VLMs specifically; the Qwen3-VL~8B comparison, with the lowest failure rate, is least confounded.}

\begin{table}[tb]
\caption{Latency and peak GPU memory, one laptop, cropped/multiple-choice, $n{=}40$/model. Latency is right-skewed for every VLM, tracking the CPU-spillover pattern noted above.}
\label{tab:latency}
\centering
\footnotesize
\renewcommand{\arraystretch}{0.9}
\begin{tabular}{@{}lccccc@{}}
\toprule
Model (params) & Mean & Median & p90 & Peak GPU mem. \\
\midrule
BioCLIP (300M) & 0.57s & 0.56s & 0.58s & N/A$^\dagger$ \\
Gemma3 4B & 1.99s & 1.76s & 1.95s & 4.1 GB \\
Qwen3-VL 2B & 19.40s & 6.38s & 50.11s & 4.1 GB \\
Qwen3-VL 4B & 9.74s & 4.86s & 28.22s & 5.6 GB \\
Qwen3-VL 8B & 30.04s & 18.02s & 103.29s & 7.2 GB \\
\bottomrule
\end{tabular}

{\footnotesize $^\dagger$Ran on CPU here, not GPU; footprint would be far below the VLMs' even if GPU-resident, given 300M params.}
\end{table}

\revise{Latency is highly right-skewed for every VLM (median well below mean -- e.g.\ Qwen3-VL~8B: 18.0s median vs.\ 30.0s mean, p90 103.3s), consistent with the CPU-spillover pattern noted above; the same effect appears, more mildly, at 4B and even 2B. Qwen3-VL~2B's mean latency (19.4s) exceeds 4B's (9.7s) despite being the smaller model, echoing the accuracy non-monotonicity discussed in RQ2 -- since that finding dissolved under a larger sample, we treat this latency ordering with the same caution rather than as established at this sample size ($n{=}40$). Peak memory tracks quantized weight size plus context/KV-cache overhead rather than parameter count alone: Gemma3~4B and Qwen3-VL~2B land at nearly identical footprints (4.1GB) despite different architectures and size.}

\begin{table}[tb]
\caption{Species accuracy by image source, multiple-choice prompting, 95\% CI on the gap; 200-image sample for every model.}
\label{tab:domain-gap}
\centering
\small
\begin{tabular}{@{}lccc@{}}
\toprule
Model & Clean & Trap & Gap (95\% CI) \\
\midrule
Qwen3-VL 2B & 39.2\% & 22.8\% & 16.3 [11.2, 21.5] \\
Qwen3-VL 4B & 36.7\% & 25.2\% & 11.5 [6.3, 16.7] \\
Qwen3-VL 8B & 56.5\% & 34.2\% & 22.3 [16.8, 27.8] \\
Gemma3 4B   & 30.5\% & 18.2\% & 12.3 [7.5, 17.1] \\
BioCLIP$^\dagger$ & 89.0\% & 71.0\% & 18.0 [10.4, 25.6] \\
\bottomrule
\end{tabular}

{\footnotesize $^\dagger$Cropped treatment only, the sole treatment run in both domains for BioCLIP.}
\end{table}

\noindent \textbf{RQ2: family, prompting, and treatment, within the 2--8B range.} Scaling within Qwen3-VL is not simply monotonic with more data: the original 100-image sample suggested 4B trailed 2B, but on the expanded 200-image sample this gap is statistical noise (30.9\% vs.\ 31.0\%, 95\% CI [$-3.6$, 3.8]) -- a useful illustration of why replication checks matter, since a real-looking $n{=}100$ effect did not survive $n{=}200$. The 8B variant's advantage held and grew clearer (45.3\%). We stress this non-monotonicity is established only within 2--8B; whether it persists at 32B or larger Qwen3-VL variants is an open question outside our scope. At matched parameter count, Qwen3-VL~4B still exceeds Gemma3~4B (30.9\% vs.\ 24.3\%). \emph{Which} prompting style wins does not replicate across evaluation sets for 3 of 4 models -- a genuinely different, unstable pattern versus the scaling/family effects above. Image treatment shows a modest, consistent effect: \emph{cropped} is the lowest-accuracy treatment on both sets (31.7\% vs.\ 33.9\%/33.1\% for original/boxed, pooled across all four VLMs at $n{=}1{,}600$/treatment), plausibly because tight cropping removes contextual scene cues -- a real but secondary driver next to the domain gap above.

\noindent \textbf{RQ3: what governs accuracy.} Every model shows a substantial clean-to-trap accuracy drop (\cref{tab:domain-gap}), 11.5--22.3 points for VLMs and 18.0 for BioCLIP, replicating on the focus set (multiple-choice: 9.6--26.2 pts; open-set: 17.2--26.6 pts; all significant at 95\%) and holding at species/genus/family level alike -- indicating general image legibility (blur, low light, partial framing), not fine-grained discrimination, drives the drop. Critically, \textbf{BioCLIP's own gap (18.0 [10.4, 25.6]) is statistically indistinguishable from the best VLM's (22.3 [16.8, 27.8])}: even a specialist trained on curated biological imagery degrades comparably under field conditions, so this degradation is unlikely to be a general-purpose-model weakness -- it more plausibly reflects the domain shift itself. Taxonomically, accuracy increases monotonically from species (37--44\%) to order level (77--80\%) on both sets: models are far more often in the right general area than exactly correct. The hardest species fail via two distinct mechanisms: confusion with a specific taxonomic lookalike (e.g.\ \emph{Odocoileus hemionus} with its congener, 3.1\% accuracy) versus near-absent recognition of a rare, distinctive species with no clear confusion partner (\emph{Orycteropus afer}, 17.1\%, errors scattered across unrelated taxa).

\textbf{Hallucination and candidate-list use.}
\label{sec:hallucination}
Under open-set prompting, 5.9--9.6\% of responses are syntactically valid but taxonomically nonexistent (broad set $n{=}1{,}949$: 90.0\% real species/wrong ID, 6.3\% real genus/invalid species, 3.3\% fabricated genus). The \emph{ranking} of fabrication rate across models is identical on both evaluation sets (Gemma3~4B fabricates 3--8$\times$ more than any Qwen3-VL variant; rate decreases monotonically with Qwen3-VL scale within the range tested) -- a materially more robust finding than either point estimate alone. Separately, 19.3\% of broad-set multiple-choice predictions named one of the 41 species never actually drawn as ground truth, evidence models engage the full 96-species list rather than a smaller effective subset.

\section{Discussion, Limitations, and Conclusion}
\label{sec:discussion}

Edge-deployable VLMs (2--8B) possess substantial, far-above-chance taxonomic knowledge -- but fall well short of a domain-specific specialist, degrade sharply and near-uniformly under field conditions, and under open-ended querying can confidently generate species names that do not exist. The relative severity of this fabrication risk across models is a robust, replicated result even where other comparisons (prompt-format preference) are not. Three findings reframe how this should be understood. First, BioCLIP's specialist advantage holding despite its far smaller parameter count (300M vs.\ up to 8B) indicates the gap traces to specialized training data, not raw model scale -- a claim that does not require, and is not weakened by, leaving frontier-scale VLMs untested. Second, BioCLIP's own domain gap being statistically indistinguishable from the best VLM's indicates the clean-to-field accuracy drop is a property of image degradation itself, not a gap specific to general-purpose pretraining -- closing it likely requires better field-image robustness broadly, not simply more biological training data. Third, the Qwen3-VL 2B-vs-4B scaling reversal that dissolved under a doubled sample directly demonstrates why this paper's two-evaluation-set, replication-first methodology matters: a single sampled evaluation set can manufacture findings a second, independent sample does not confirm.

\textbf{Limitations.} All VLM results use Q4-quantized weights, the common configuration for edge deployment but not full precision; \revise{because BioCLIP is not quantized, part of the VLM-vs-BioCLIP gap in \cref{tab:bioclip} is confounded with this precision difference rather than being purely a training-data effect. BioCLIP's forced-choice head also cannot fail to produce an answer the way a generative VLM's free-form response can (RQ1); the gap reflects both taxonomic knowledge and instruction-following reliability, not fully disentangled here. Our broad and focus sets are sampled from the same underlying pool rather than disjoint camera networks, so replication demonstrates robustness to resampling, not generalization to a new site. We report BioCLIP's domain gap as statistically indistinguishable from the best VLM's based on overlapping 95\% CIs, not a formal difference test; BioCLIP's own interval is wide, so this is failure to detect a difference, not proof of equivalence. Every claim is scoped to the 2--8B range tested -- whether these patterns persist at 32B-plus scale, a full-precision ablation, and a frontier-scale ceiling reference are left to future work.}

Taken together, these results suggest edge-deployable VLMs currently hold breadth of taxonomic knowledge without the reliability required for unsupervised ecological deployment; closed-set prompting against a known candidate list is markedly safer than open-ended generation for this reason, independent of which format happens to score higher on any given sample.

\bibliographystyle{splncs04}
\bibliography{main}

\end{document}